\documentclass{article}
\usepackage{ijcai23}

\usepackage{times}
\usepackage{soul}
\usepackage{url}
\usepackage[utf8]{inputenc}
\usepackage[small]{caption}
\usepackage{graphicx}
\usepackage{amsmath}
\usepackage{amssymb}
\usepackage{multirow}
\usepackage{amsthm}
\usepackage{booktabs}
\usepackage{algorithm}
\usepackage{algorithmic}
\usepackage[switch]{lineno}
\usepackage[hidelinks]{hyperref}

\title{Orientation-Independent Chinese Text Recognition in Scene Images}

\author{
Haiyang Yu, 
Xiaocong Wang, 
Bin Li\thanks{Corresponding author}, 
Xiangyang Xue$^*$
\affiliations
Shanghai Key Laboratory of Intelligent Information Processing\\
School of Computer Science, Fudan University\\
\emails
\{hyyu20, xcwang20, libin, xyxue\}@fudan.edu.cn
}
\begin{document}

\maketitle

\begin{abstract}
    Scene text recognition (STR) has attracted much attention due to its broad applications. The previous works pay more attention to dealing with the recognition of Latin text images with complex backgrounds by introducing language models or other auxiliary networks. Different from Latin texts, many vertical Chinese texts exist in natural scenes, which brings difficulties to current state-of-the-art STR methods. In this paper, we take the first attempt to extract orientation-independent visual features by disentangling content and orientation information of text images, thus recognizing both horizontal and vertical texts robustly in natural scenes. Specifically, we introduce a Character Image Reconstruction Network (CIRN) to recover corresponding printed character images with disentangled content and orientation information. We conduct experiments on a scene dataset for benchmarking Chinese text recognition, and the results demonstrate that the proposed method can indeed improve performance through disentangling content and orientation information. To further validate the effectiveness of our method, we additionally collect a Vertical Chinese Text Recognition (VCTR) dataset. The experimental results show that the proposed method achieves 45.63\% improvement on VCTR when introducing CIRN to the baseline model.
\end{abstract}

\begin{figure}[t]
    \centering
    \includegraphics[width=0.45\textwidth]{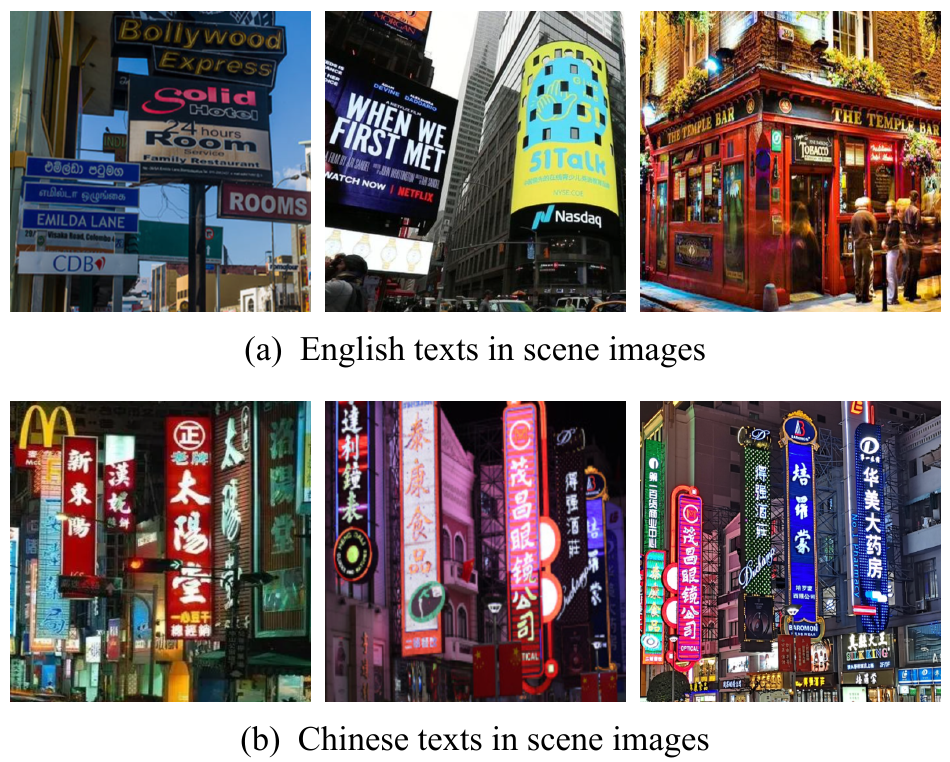}
    \caption{In street-view images, most English texts are horizontal; in contrast, vertical Chinese texts are commonly-seen as well. }
    \label{fig:1}
\end{figure}

\section{Introduction}
\label{intro}
Scene text recognition (STR) has received much attention in the field of computer vision due to its broad range of applications, such as traffic sign recognition~\cite{sermanet2011traffic} and text image retrieval~\cite{wang2019camp}. It aims to transcribe texts from natural images into sequences of digital characters. Reading texts from natural images faces many difficulties, \textit{e.g.}, text distortion, partial occlusion, and complex backgrounds. Different from Latin text recognition, Chinese text recognition poses additional challenges, such as commonly-seen vertical texts and complicated sequential patterns~\cite{chen2021benchmarking}. These unique features make Chinese text recognition a challenging task. 

Compared with Latin texts, Chinese texts are more likely to appear in the vertical orientation due to the commonly-used traditional couplets or signboards in natural scenes (as shown in Figure~\ref{fig:1}(b)). On the contrary, there are few vertical Latin texts due to different inherent reading habits (as shown in Figure~\ref{fig:1}(a)). Most of the early methods~\cite{yin2017scene,liu2018char,shi2016end} are specially designed for Latin text recognition, and limited to horizontal texts. Thus, they can hardly handle text instances with various shapes such as curved and vertical texts, leading to a serious impact on recognizing Chinese texts in scene images. To tackle curved texts, some methods~\cite{shi2018aster,li2019show} introduce a rectification network~\cite{jaderberg2015spatial} to straighten irregular text instances or rely on the 2D attention mechanism to locate each character. In addition, researchers have tried to introduce linguistic knowledge and corpora to improve the performance on curved texts~\cite{fang2021read,yu2020towards}. However, these methods are still inefficient for vertical text recognition since the layout of vertical texts is completely different from horizontal or curved texts. Some Chinese character recognition methods~\cite{wu2019joint} have attempted to improve the robustness of models for rotated characters, but they cannot be directly applied to text line recognition. On the whole, existing scene text recognition methods still have difficulties in dealing with vertical Chinese texts. Thus, developing a network to learn visual features that are independent of text orientation is crucial for recognizing vertical Chinese texts.

We observe that visual features contain not only the content information that determines character predictions but also text orientation information. Therefore, in this paper, we try to disentangle the content and orientation information from the visual features to obtain orientation-independent features for accurate recognition of vertical Chinese texts. The proposed method consists of a customized ResNet~\cite{he2016deep} encoder, a transformer-based decoder~\cite{vaswani2017attention}, and a character image reconstruction network. By making modifications to ResNet, the encoder captures more details and preserves more visual features. The character image reconstruction network contains a content information extractor, an orientation information extractor, and a reconstruction module. The content information extractor is used to obtain content information from visual features, and the orientation information extractor disentangles orientation information. We decouple the content and orientation information of horizontal and rotated vertical characters and exchange their orientation information to reconstruct corresponding printed character images. Finally, we use a transformer-based decoder to capture the semantic dependencies between characters to generate final predictions.

To benchmark the performance of existing state-of-the-art methods in vertical Chinese text recognition, we collect a vertical Chinese text recognition (VCTR) dataset from PosterErase~\cite{postererase}. The experimental results show that our method outperforms existing STR models by a large margin on VCTR. In addition, we achieve better results on a generic Chinese text recognition dataset. The code of our method and VCTR dataset are available at GitHub\footnote{https://github.com/FudanVI/FudanOCR/orientation-independent-CTR}. The contributions of this paper can be summarized as follows:
 \begin{itemize}
    \item We collect a Vertical Chinese Text Recognition (VCTR) dataset to benchmark the performance of vertical Chinese text recognition since vertical texts are the key issue affecting Chinese scene text recognition.
    \item We take the first attempt to disentangle the content and orientation information from visual features with a character image reconstruction network, which can eliminate the disturbance of text orientation.
    \item  The proposed method significantly outperforms the existing methods on vertical Chinese text recognition and also achieves new state-of-the-art results on a Chinese scene text recognition dataset. 
\end{itemize}

\section{Related Work}
Scene text recognition (STR) has been a long-standing research topic in computer vision. Early works in this field focus on utilizing low-level features such as histograms of oriented gradient descriptors~\cite{KaiWang2011EndtoendST}, connected components~\cite{LukasNeumann2012RealtimeST}, and so on. With the rapid development of deep learning, STR research has made significant progress in the last few years. Based on their linguistic categories, we divide them into two categories: Latin text recognition and Chinese text recognition.

\subsection{Latin Text Recognition}
Latin scene text recognition can be divided into two categories: regular and irregular text recognition. The sequence-to-sequence models based on the CTC loss~\cite{graves2006connectionist,shi2016end} and attention mechanism~\cite{cheng2017focusing} have made great progress in regular text recognition. However, these methods struggle to handle curved or rotated texts. For irregular texts, previous methods~\cite{zhan2019esir,MingkunYang2019SymmetryConstrainedRN,shi2018aster} tend to integrate a spatial transformer module into an attention-based framework to rectify the curved text images to the horizontal form, but the predefined transformation space limits their generalization capabilities. The segmentation-based methods~\cite{liao2019scene,ZhaoyiWan2020TextScannerRC} first detect characters and then integrate characters into text predictions. Some recently proposed approaches attempt to use linguistic rules to aid the recognition process, showing strong performance on irregular text recognition. For example, ABINet~\cite{fang2021read} and VisionLAN~\cite{wang2021two} develop a specific module to integrate language information into text recognition. The aforementioned approaches are all specially designed for Latin text recognition, and cannot work well when facing Chinese text recognition due to the large alphabet and commonly-seen vertical texts.

\subsection{Chinese Text Recognition}
Due to complex inner structures of Chinese characters, some methods~\cite{yu2022chinese,zu2022chinese} are proposed to recognize Chinese characters. DenseRAN~\cite{WenchaoWang2018DenseRANFO} treats a Chinese character as a composition of two-dimensional structures and radicals. Based on DenseRAN, STN-DenseRAN~\cite{wu2019joint} further employs a rectification block to handle distorted character images. HDE~\cite{cao2020zero} designs a unique embedding vector for each Chinese character according to its radical-level constitution. In~\cite{JingyeChen2021ZeroShotCC}, characters are decomposed into a combination of five strokes in Chinese, and the predicted stroke sequence is transformed to a specific character through a matching-based strategy. Recently, some works~\cite{chen2021benchmarking,su2023privacy} focus on Chinese text recognition (CTR). For instance, the authors of ~\cite{chen2021benchmarking} proposed a benchmark for CTR and introduced radical-level supervision to improve the performance of text recognition models on CTR. SVTR~\cite{du2022svtr} proposes a transformer-based framework, utilizing global mixing and local mixing to perceive the inter-character and intra-character patterns, respectively. It performs well on the Chinese scene dataset. However, these approaches mainly focus on Chinese character or horizontal text recognition, while ignoring the commonly-seen vertical texts. 

% We try to make the recognizer more robust to vertical text images by building a character reconstruction network to obtain orientation-independent text content.

\begin{figure*}[t]
    \centering
    \includegraphics[width=1.0\textwidth]{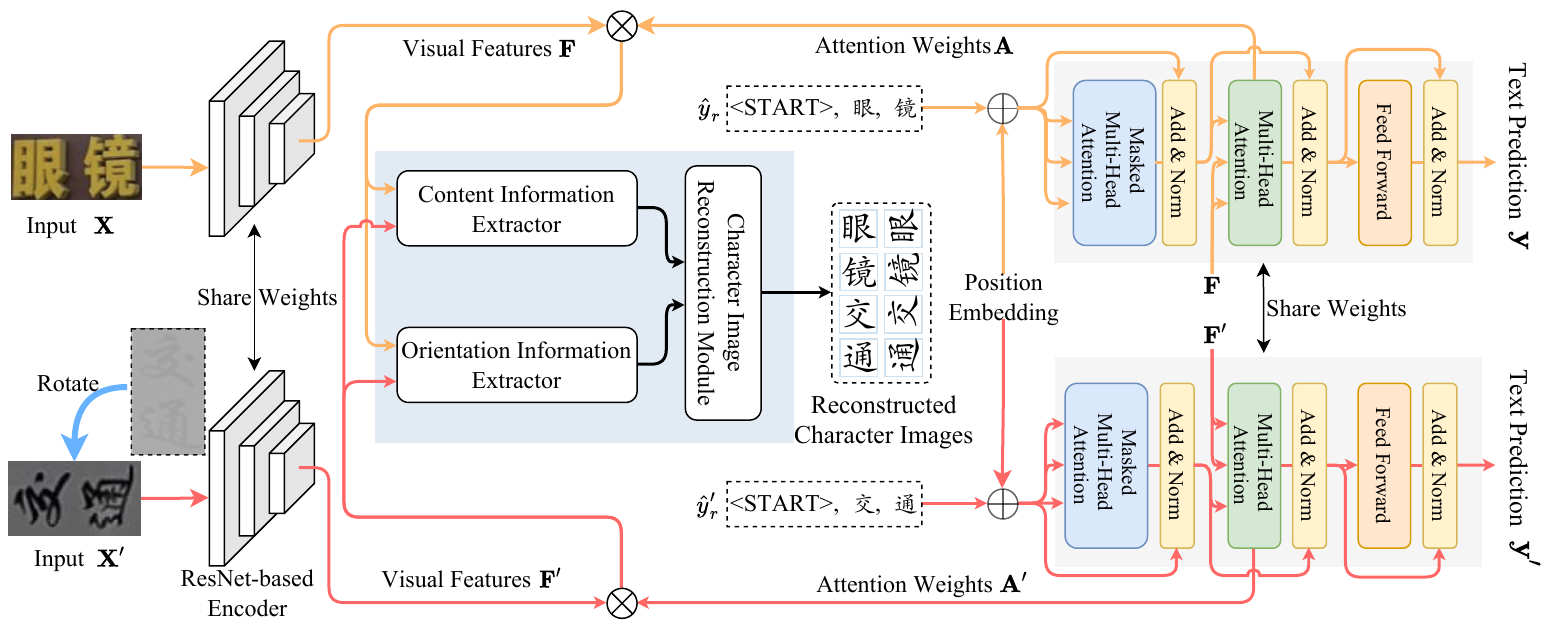}
    \caption{The overall architecture of the proposed method. It consists of a ResNet-based encoder, a transformer-based decoder, and a character image reconstruction network. The data flow of horizontal and vertical text images is in orange and red lines, respectively. The blue box represents the character image reconstruction network; the grey boxes represent the transformer-based decoder.}
    \label{fig:framework}
\end{figure*}

\section{Methodology}
In this section, we first review the commonly-used encoder-decoder framework in scene text recognition. Then, we analyze the information contained in extracted visual features by the encoder. Finally, we introduce the details of each module in the proposed architecture.

\subsection{Generic Framework}
In the past few years, researchers tend to adopt the encoder-decoder framework to solve the text recognition task. Generally, a ResNet-based backbone~\cite{he2016deep} is employed as the encoder to extract visual features $\textbf{F}$. Subsequently, the features $\textbf{F}$ are fed into designed decoders, such as the attention-based decoder~\cite{li2019show} and the transformer-based decoder~\cite{vaswani2017attention}. Both of these decoders are composed of two modules: the attention module and the prediction module. At the $t$-th time step, the attention module calculates the glimpse vector $\mathbf{g}_t$ as follows:
\begin{equation}
    \label{eq1}
    \mathbf{g}_t = \sum_{ij} \alpha^t_{ij}\mathbf{f}_{ij}
\end{equation}
where $\mathbf{f}_{ij}$ represents the feature vector at the position ($i, j$) of $\mathbf{F}$ and $\alpha^t_{ij}$ is the attention weight of $\mathbf{f}_{ij}$ at the $t$-th time step. Finally, the glimpse vector $\mathbf{g}_t$ is taken as the input of the prediction module to predict the corresponding character or the end token $\langle$EOS$\rangle$:

\begin{equation}
    \label{eq2}
    \mathbf{y}_t = \text{Softmax}(\mathbf{W}\mathbf{g}_t+b)
\end{equation}
where $\mathbf{W}$ and $b$ represent the linear transformation and the bias of the prediction module, respectively.

\subsection{Visual Features Dissection}
The existing methods, which are mostly designed for tackling Latin text recognition, rarely consider the problem of vertical text recognition. In contrast, vertical Chinese texts are commonly-seen in natural scenes, as shown in Figure~\ref{fig:1}. To recognize vertical and horizontal texts simultaneously, some researchers~\cite{li2019show} proposed to rotate those images with height larger than width by 90 degrees anticlockwise and then feed them into recognition models. The experimental results demonstrate that this strategy can indeed improve the performance on vertical text images. However, this strategy will make recognizers confused to a certain degree since recognizers are forced to classify completely different visual features (\textit{i.e.}, the visual features of horizontal and rotated vertical characters) into the same character. Through experiments, we observe that when a small proportion of vertical text images are included in the training set, the performance of recognizers will descend when this strategy is used at the training stage.

% According to Equation~\ref{eq1} and~\ref{eq2}, the predicted characters only depend on the aggregated visual features, \textit{i.e.}, the glimpse vector $\mathbf{g}_t$. Therefore, the rotation invariance of recognizers mostly stem from the max pooling layers in the encoder when using the aforementioned strategy at the training stage. 

Based on the above observation, we speculate that the extracted visual features contain not only content information but also orientation information of characters. To verify the conjecture about orientation information, we calculate the similarity $S_o$ between the visual features of two different text images with the same orientation and the similarity $S_c$ between two text images with different orientations but with the same characters. We observe that $S_o$ tends to be larger than $S_c$, implying that the visual features indeed contain the orientation information. Therefore, in this paper, we try to develop a character image reconstruction network to disentangle the orientation-independent content information from the visual features, which will make the recognizer more robust to vertical text images.

\subsection{Proposed Method} 
\paragraph{Overview.} As depicted above, we observe that the extracted visual features contain not only the content information, which determines the character prediction, but also the orientation information, which is useless for final predictions. Thus, we attempt to disentangle the content and orientation information from the extracted visual features. As shown in Figure~\ref{fig:framework}, we take TransOCR~\cite{chen2021benchmarking} as the baseline model, which consists of a ResNet-based encoder and a transformer-based decoder. In this paper, we additionally develop a Character Image Reconstruction Network (CIRN). In the following, we introduce the details of our method.

\paragraph{ResNet-based Encoder.} Given the input text image $\mathbf{X}$, the ResNet-based encoder is employed to extract its visual features $\mathbf{F} \in \mathbb{R}^{\frac{H}{8} \times \frac{W}{8} \times C}$ ($H$ and $W$ represent the height and width of the input image $\mathbf{X}$, respectively). We adopt ResNet-34 as the main body of the encoder and modify some layers in the original ResNet-34. First, we replace the $7\times7$ kernel of the first convolution layer with a $3\times3$ kernel since the smaller kernel can capture more local details for recognizing text images. In addition, we drop the last convolution block to reduce the number of parameters in the encoder and improve the efficiency of feature extraction. Finally, we remove the max pooling layer of the third convolution block in ResNet-34 to reserve more visual features for the subsequent decoder. Although the max pooling layer can make the model have rotation invariance to some extent~\cite{laptev2016ti}, the loss of visual information outweighs the gains of rotation invariance. 

\paragraph{Transformer-based Decoder.} As shown in Figure~\ref{fig:framework}, the transformer-based decoder consists of three modules: the masked multi-head attention module, the multi-head attention module and the feed forward module. The masked multi-head attention module, taking the right-shifted ground truth $\hat{y}_r$ as input, is to capture the semantic dependence between characters. Then, the multi-head attention module calculates the attention weights $\mathbf{A}$ between the extracted visual features $\mathbf{F}$ and $\hat{y}_r$. Finally, the weighted features are input to the feed forward module to extract deeper features, which is used to generate the final predictions $\mathbf{y}$ through a linear layer.

\paragraph{Character Image Reconstruction Network.} To disentangle the content information from $\mathbf{F}$ and avoid the disturbance of orientation information, we propose a Character Image Reconstruction Network (CIRN). Specifically, we first obtain the visual features $\mathbf{F}_c$ of each character in the text image by a position-wise multiplication between $\mathbf{F}$ extracted by the encoder and the attention weights $\mathbf{A}$ from the transformer-based decoder. Then, $\mathbf{F}_c$ is fed into CIRN to disentangle its content and orientation information. 

The proposed CIRN contains a content information extractor, an orientation information extractor and a reconstruction module. The content information extractor simply adopts a $1\times1$ convolution layer rather than other complex structures (more analysis is shown in Section~\ref{dis}). The orientation information extractor additionally employs a global average pooling layer after a $1\times1$ convolution layer to extract the orientation information since the orientation information should be obtained from a global perspective. Among the sampled data in a training batch, there are horizontal and vertical text images. Assuming that the orientation and content information of a horizontal character $a$ and a rotated vertical character $b$ are denoted as $O_a$, $O_b$, $C_a$, and $C_b$, we exchange their orientation information to reconstruct corresponding printed character images. Formally, four character representations can be obtained as follows:
\begin{equation}
    H_a = \text{Fuse}(C_a, O_a), V_a = \text{Fuse}(C_a, O_b)
\end{equation}
\begin{equation}
    H_b = \text{Fuse}(C_b, O_a), V_b = \text{Fuse}(C_b, O_b)
\end{equation}
where ``Fuse'' indicates that we concatenate the vector of orientation information to each position of the content information. Then, four character representations are fed into the reconstruction module to generate corresponding printed horizontal character images $H$ and rotated vertical character images $V$. Specifically, the reconstruction module simply consists of 5 layers of deconvolution with the kernel size of 5 and the stride of 2.

\paragraph{Loss Functions.} Four loss functions are introduced to supervise the proposed method:

1) Text prediction loss $\mathcal{L}_{t}$, supervising the prediction of texts in input images, is calculated by:
\begin{equation}
    \mathcal{L}_{t} = \text{CE}(\mathbf{y}, \hat{\mathbf{y}})
\end{equation}
where ``CE'' represents the cross entropy loss function; $\hat{\mathbf{y}}$ denotes the ground truth of predicted texts.

2) Orientation classification loss $\mathcal{L}_{o}$ is used to supervise the prediction of character orientation, which can be regarded as a binary classification problem and be computed as follows:
\begin{equation}
    \mathcal{L}_{o} = \text{CE}(O, \hat{O})
\end{equation}
where $O$ and $\hat{O}$ denote the orientation prediction of characters and corresponding ground truth respectively. $\hat{O}$ can be obtained by comparing input images' height and width.

3) Content classification loss $\mathcal{L}_{c}$, similar to $\mathcal{L}_{o}$, is calculated by:
\begin{equation}
    \mathcal{L}_{c} = \text{CE}(C, \hat{C})
\end{equation}
where $C$ represents the character predictions through the content information, and $\hat{C}$ denotes corresponding ground truth.

4) Character image reconstruction loss $\mathcal{L}_{r}$ is employed to constrain the reconstruction of horizontal and rotated vertical character images, which can be formulated as:
\begin{equation}
    \mathcal{L}_{r} = \text{MSE}(H, \hat{H}) + \text{MSE}(V, \hat{V})
\end{equation}
where $\hat{H}$ and $\hat{V}$ represent the printed character images in font Simsun corresponding to $H$ and $V$.

Therefore, the overall loss function $\mathcal{L}$ of the proposed method is the weighted sum of the above four loss functions:
\begin{equation}
    \mathcal{L} = \mathcal{L}_{t} + \alpha \mathcal{L}_o + \beta \mathcal{L}_c + \gamma \mathcal{L}_r
\end{equation}
where $\alpha, \beta,$ and $\gamma$ are hyperparameters to balance these four loss functions.

\section{Experiments}
In this section, we first introduce the details of the adopted scene dataset in~\cite{chen2021benchmarking} and the collected Vertical Chinese Text Recognition (VCTR) dataset. Then, we introduce the utilized evaluation methods and the implementation details of our method. Finally, we present the results of ablation studies and experiments.

\begin{figure}[t]
    \centering
    \includegraphics[width=0.47\textwidth]{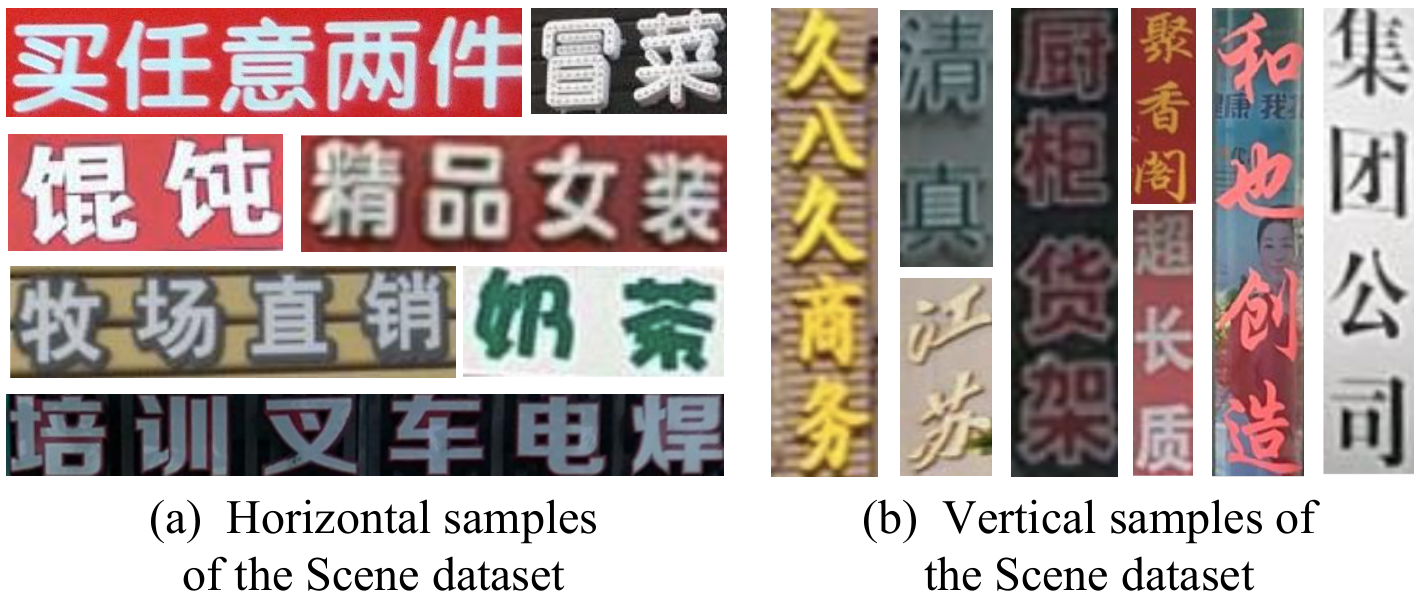}
    \caption{Some examples of horizontal and vertical text images in the scene dataset.}
    \label{fig:scene}
\end{figure}

\begin{figure}[t]
    \centering
    \includegraphics[width=0.47\textwidth]{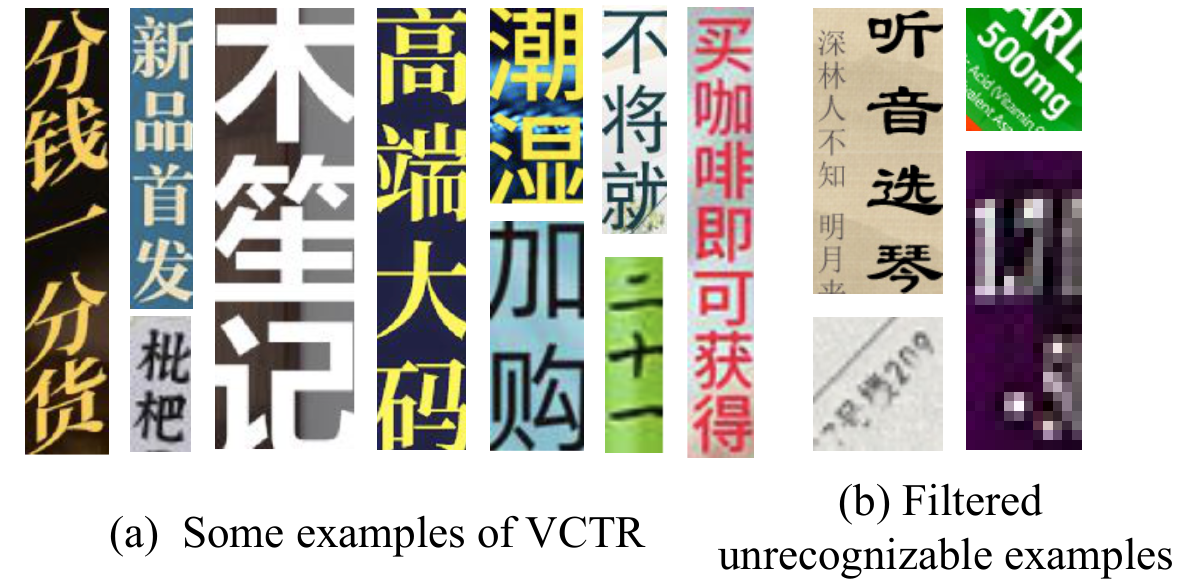}
    \caption{Some examples of VCTR are shown on the left. Only vertical texts are reserved for VCTR. Some unrecognizable examples (\textit{e.g.}, multi-line and oblique texts) that are removed during collecting VCTR are shown on the right.}
    \label{fig:VCTR}
\end{figure}

\subsection{Datasets}
\paragraph{Scene dataset.} The scene dataset is collected by~\cite{chen2021benchmarking} and derives from six existing datasets, including RCTW~\cite{shi2017icdar2017}, ReCTS~\cite{zhang2019icdar}, LSVT~\cite{sun2019icdar}, ArT~\cite{chng2019icdar2019}, and CTW~\cite{yuan2019large}. This dataset contains 509,164 samples for training, 63,645 for validation, and 63,646 for test. Some examples of this dataset are shown in Figure~\ref{fig:scene}.

\paragraph{VCTR.} To validate the effectiveness of our method in tackling vertical text images, we collect a Vertical Chinese Text Recognition (VCTR) dataset from PosterErase~\cite{postererase}, which is originally proposed for the scene text erasing task. PosterErase includes 58,114 training samples, 148 validation samples and 146 test samples. We only collect vertical text images from the training set of PosterErase since its annotations contain the orientation information of cropped text areas. We obtain the VCTR dataset through the following steps: 1) Filter out the cropped text areas annotated as horizontal ones and reserve the remaining vertical text areas; 2) Remove those unrecognizable text areas (some examples are shown in Figure~\ref{fig:VCTR}(b)). 3) Annotate the reserved text areas. Finally, we collect 5,456 samples for VCTR to verify the effectiveness of our method. Some examples of this dataset are shown in Figure~\ref{fig:VCTR}(a).

% 可以补充一个字符比例的图

\subsection{Evaluation Methods}
We follow~\cite{chen2021benchmarking} to utilize four rules to convert the predictions and labels: 1) Convert full-width characters to half-width characters; 2) Convert traditional Chinese characters to simplified characters; 3) Convert uppercase letters to lowercase letters; 4) Remove all spaces. After transforming the predictions and labels, two mainstream metrics are employed to evaluate our method: Accuracy (ACC) and Normalized Edit Distance (NED). The ACC is calculated as follows:
\begin{equation}
    \text{ACC} = \frac{1}{N} \sum^N_{i=1}\mathbb{I}(y_i=\hat{y}_i)
\end{equation}
where $y_i$ and $\hat{y}_i$ denote the $i$-th transformed prediction and label, respectively; $N$ is the number of text images; $\mathbb{I}$($\cdot$) denotes the indication function. The NED is computed by:
\begin{equation}
    \text{NED} = 1 - \frac{1}{N} \sum^N_{i=1} \text{ED}(y_i, \hat{y}_i)/\text{maxlen}(y_i, \hat{y}_i)
\end{equation}
where ``ED'' and ``maxlen'' represent the edit distance and the maximum sequence length, respectively.

\begin{table}[t]
\renewcommand{\arraystretch}{1.0}
\small
\centering
\scalebox{1.0}{\begin{tabular}{cccccc}
\toprule
Method & Rotation & $\mathcal{L}_c$ & $\mathcal{L}_o$ & $\mathcal{L}_r$ & ACC / NED \\
\midrule
TransOCR (base) & & & & & 67.98 / 0.815 \\
Ours & \checkmark & & & & 69.54 / 0.836\\
Ours & \checkmark &  \checkmark & \checkmark & & 69.96 / 0.841\\
Ours & \checkmark & \checkmark  & & \checkmark & 71.53 / 0.850\\
Ours & \checkmark  &  & \checkmark & \checkmark & 72.08 / 0.853\\
Ours (final) & \checkmark & \checkmark & \checkmark & \checkmark & \textbf{73.17} / \textbf{0.865} \\
\bottomrule
\end{tabular}
}
\caption{The ablation studies of our method on the validation set of the scene dataset. All the training settings of these methods (\textit{e.g.}, the training data) are the same. The column ``Rotation'' denotes that whether the rotation strategy is utilized at the training stage. $\mathcal{L}_c$, $\mathcal{L}_o$, and $\mathcal{L}_r$ represent whether the content classification loss, the orientation classification loss, and the character image reconstruction loss are used for supervision, respectively.}
\label{tab:ablation}
\end{table}

\begin{figure*}[t]
    \centering
    \includegraphics[width=0.92\textwidth]{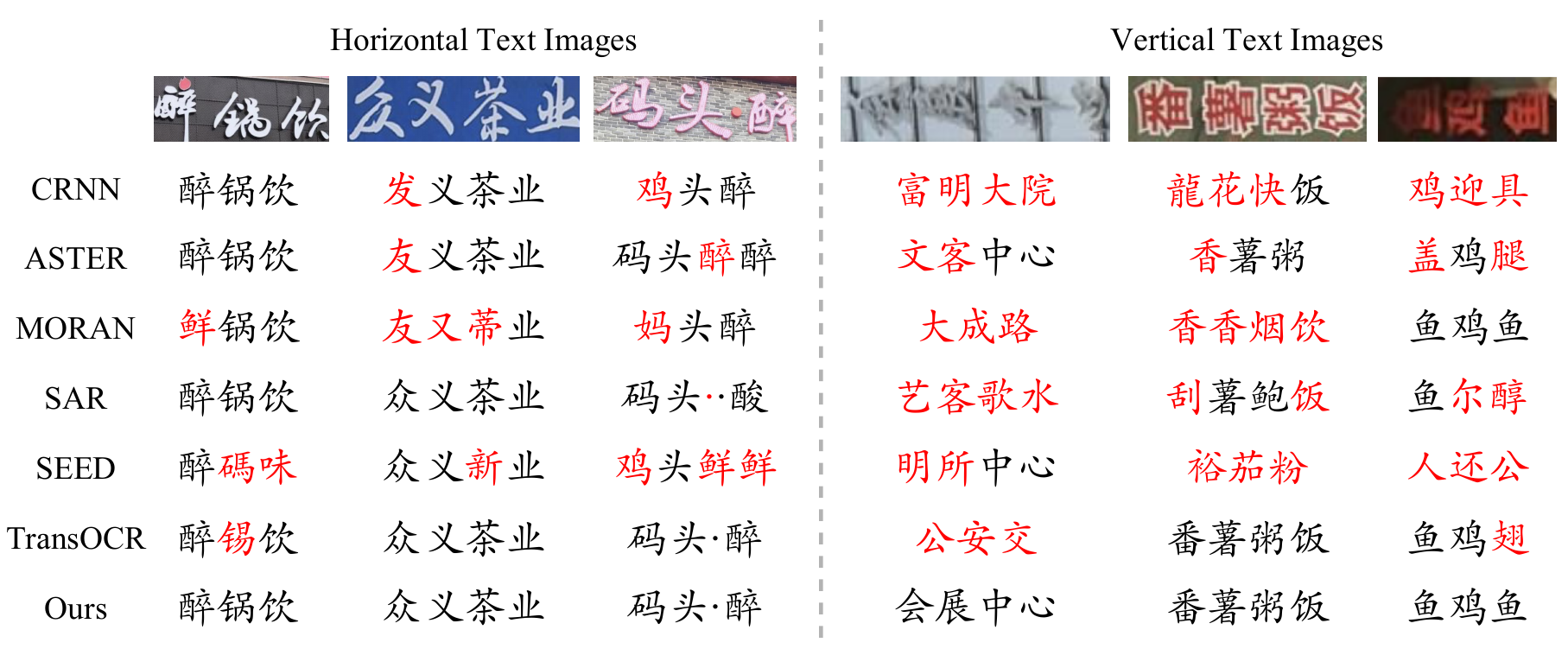}
    \caption{Comparison of recognition results on the scene dataset. The vertical text images are rotated by 90 degrees anticlockwise for convenience. The proposed method performs well on both the horizontal and vertical texts. The characters in red are wrongly predicted.}
    \label{fig:res_scene}
\end{figure*}

\begin{figure}[t]
    \centering
    \includegraphics[width=0.50\textwidth]{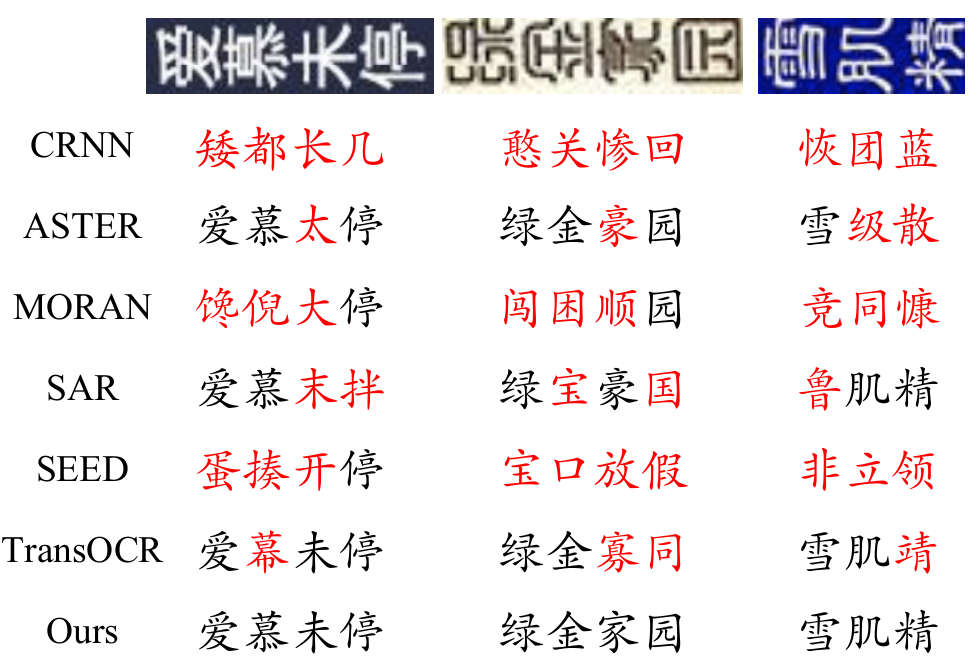}
    \caption{Comparison of recognition results on VCTR. The shown samples are rotated by 90 degrees anticlockwise. }
    \label{fig:res_VCTR}
\end{figure}

\subsection{Implementation Details}
Our method is implemented with PyTorch, and all experiments are conducted on an NVIDIA RTX 2080Ti GPU with 11GB memory. The AdaDelta~\cite{zeiler2012adadelta} optimizer is adopted to train our model with an initial learning rate 1.0, and the hyperparameters $\rho$ and weight decay are set to 0.9 and $10^{-4}$, respectively. The batch size is set to 64. For fair comparison with the previous method~\cite{chen2021benchmarking}, the input text images are resized into $32\times256$. For vertical text images, we follow~\cite{li2019show} to rotate them by 90 degrees anti-clockwise. However, different from the rule in~\cite{li2019show} that regards the samples with height larger than width as vertical ones, we assume that the samples with height larger than 1.5$\times$ width are vertical text images.

\subsection{Ablation Study}
To disentangle the content and orientation information from visual features, we introduce three additional loss functions to supervise the proposed CIRN. In this section, we conduct ablation studies on these key loss functions and the rotation strategy, and the experimental results are shown in Table~\ref{tab:ablation}. All methods are trained on the training set of the scene dataset. Through experimental results, we observe that introducing the rotation strategy and two classification losses (\textit{i.e.}, $\mathcal{L}_c$ and $\mathcal{L}_o$) can achieve 1.56\% and 0.42\% improvement, respectively. When adding the character image reconstruction loss $\mathcal{L}_r$, the content and orientation information can be better disentangled and our method achieves the best performance.

\begin{table}[t]
\renewcommand{\arraystretch}{1.1}
% \small

\centering
\scalebox{0.9}{\begin{tabular}{p{0.5cm} p{0.5cm} p{0.5cm} p{2.0cm}}
\toprule
\multicolumn{3}{c}{Hyperparameter} & \multirow{2}*{ACC / NED} \\
\cmidrule{1-3}
$\alpha$ & $\beta$ & $\gamma$ & ~ \\
\midrule
1 & 1 & 1 & 70.13 / 0.832 \\
1 & 1 & 2 & 71.25 / 0.841\\
1 & 1 & 5 & \textbf{73.17} / \textbf{0.865} \\
1 & 2 & 5 & 72.31 / 0.847\\
1 & 5 & 5 & 70.76 / 0.839\\
2 & 1 & 5 & 70.30 / 0.827\\
5 & 1 & 5 & 68.15 / 0.815\\
\bottomrule
\end{tabular}
}
\caption{The experimental results of choosing appropriate hyperparameters. All results are evaluated on the validation dataset of the scene dataset.}
% When $\alpha, \beta$, and $\gamma$ are set to 1, 1, and 5 respectively, our method achieves the best performance on the validation dataset.}
\label{tab:hyperparameter}
\end{table}

\begin{table}[t]
\renewcommand{\arraystretch}{1.1}
\small

\centering
\scalebox{1.0}{\begin{tabular}{ccc}
\toprule
\multirow{2}*{Method} & \multicolumn{2}{c}{Dataset} \\
\cmidrule{2-3}
~  & Scene & VCTR \\
\midrule
CRNN~\cite{shi2016end} & 54.94 / 0.742 & 8.99 / 0.173 \\
ASTER~\cite{shi2018aster} & 59.37 / 0.801 & 19.70 / 0.434 \\
MORAN~\cite{luo2019moran} & 54.68 / 0.710 & 17.43 / 0.328  \\ 
SAR~\cite{li2019show} & 53.80 / 0.738 & 9.53 / 0.187 \\
SEED~\cite{qiao2020seed} & 45.37 / 0.708 & 8.32 / 0.193  \\
TransOCR~\cite{chen2021scene} & 67.81 / 0.817 & 18.35 / 0.341  \\
Ours & \textbf{73.29} / \textbf{0.866} & \textbf{63.98} / \textbf{0.863}  \\

\bottomrule
\end{tabular}
}

\caption{The experimental results on the test sets of the scene dataset and VCTR. ACC/NED follows the percentage format and decimal format, respectively.}
% When introducing the character image reconstruction network, the proposed method outperforms previous methods by a large margin on VCTR.}
\label{tab:whole}
\end{table}

\begin{table}[ht]
\renewcommand{\arraystretch}{1.1}
\small
\centering
\scalebox{1.0}{\begin{tabular}{ccc}
\toprule
Structure & ACC &  NED \\
\midrule
Transformer Encoder & 67.97 & 0.834\\
2$\times$ \text{Transformer Encoder} & 61.29 & 0.785\\
1$\times$1\ \text{Convolution (Ours)} & \textbf{73.17} & \textbf{0.865} \\
\bottomrule
\end{tabular}
}
\caption{Comparison of different structures adopted in the content information extractor. ``2$\times$'' indicates that two transformer layers are stacked as the structure of the content information extractor. }
\label{tab:complex}
\end{table}

\subsection{Experimental Results}
In the following experiments, our method is trained on the training set of the scene dataset in~\cite{chen2021benchmarking}. At the training stage, we evaluate the performance of our method on the validation set of the scene dataset, and reserve the optimal model to test on the test set. For all experiments, the number of heads in the transformer-based decoder is set to 4. Following~\cite{chen2021benchmarking}, we select CRNN~\cite{shi2016end}, ASTER~\cite{shi2018aster}, MORAN~\cite{luo2019moran}, SAR~\cite{li2019show}, SEED~\cite{qiao2020seed}, and TransOCR~\cite{chen2021scene} for comparison.

\paragraph{Choices of hyperparameters.} In the overall loss function, three hyperparameters are adopted to balance four introduced loss functions. We conduct experiments to choose the appropriate hyperparameters. As shown in Table~\ref{tab:hyperparameter}, when $\alpha$ and $\beta$ are set to 1, our method achieves relatively better performance. A possible reason is that the content and orientation information classification is easy to optimize in our method. Differently, our method achieves the best performance when $\gamma$ is set to 5, indicating that the character image reconstruction is crucial for disentangling the orientation and content information.

\paragraph{Experiments on the scene dataset.}
We only conduct experiments on the scene dataset in~\cite{chen2021benchmarking} since more vertical text images are contained in this scenario. The experimental results shown in Table~\ref{tab:ablation} demonstrate that the rotation strategy can alleviate the recognition of vertical text images to some extent, and achieve 1.56\% improvement in accuracy compared with the baseline model. Therefore, our method also adopts this strategy at the training and test stage. Through disentangling the content and orientation information, our method surpasses the SOTA model TransOCR by 5.48\% in accuracy, which indicates the effectiveness of our method in tackling Chinese text recognition. We visualize some recognition results of the scene dataset in Figure~\ref{fig:res_scene}. Through the visualization, we observe that the proposed method performs better on both horizontal and vertical texts. Additionally, compared with previous methods, the proposed method is more robust to artistic texts in scene images, which benefits from that the proposed CIRN implicitly pulls features of characters close to that of corresponding printed characters images.

\paragraph{Experiments on VCTR.}
To further validate the effectiveness of our method in tackling vertical text images, we also conduct experiments on the proposed VCTR dataset. Since the VCTR dataset is only used for test, our model is also trained on the scene dataset. The experimental results are shown in Table~\ref{tab:whole}. Compared with previous methods, the proposed method achieves the best performance on the VCTR dataset. Specifically, our method surpasses the SOTA method~\cite{chen2021scene} by around 45\%. Although our method is not fine-tuned on training sets similar to VCTR, it still achieves satisfying performance. Some recognition results of VCTR are shown in Figure~\ref{fig:res_VCTR}.

\section{Discussions}
\label{dis}
\paragraph{Adopting complex structures for the content information extraction.}
In the content information extractor, we adopt a simple $1\times1$ convolution layer rather than a more complex structure. In this module, we additionally try to utilize a multi-layer transformer encoder to disentangle the content information from visual features. Through the experimental results shown in Table~\ref{tab:complex}, we observe that when a more complex structure is employed in the content information extractor, the performance of our method decreases clearly, which may result from two reasons: 1) The transformer encoder contains more parameters while vertical text images account for a small share. Thus, a more complex structure for the content information extractor is hard to converge with the training of limited vertical text samples. 2) The content information can be easily disentangled from the extracted visual features with the supervision of the content classification loss $\mathcal{L}_c$ and character image reconstruction loss $\mathcal{L}_r$.
%一方面参数多样本少不好学习，另外内容信息相对容易解耦。

\paragraph{Is it better to reconstruct the whole text image?}
In this paper, we propose to reconstruct character images with different orientations, thus forcing the content information extractor to produce orientation-independent visual features. Additionally, we have attempted to reconstruct the whole text images to complete disentangling. Through experiments, however, we observe that the reconstruction loss cannot descend smoothly and the printed text images are not reconstructed well. A possible reason is that the reconstruction network cannot be aware of the position of each character. Therefore, we choose to reconstruct the printed image of each character in the input text image.

\begin{figure}[t]
    \centering
    \includegraphics[width=0.47\textwidth]{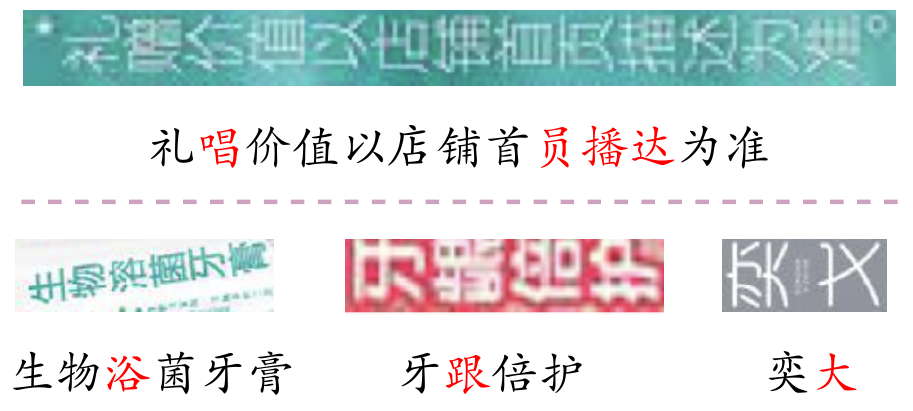}
    \caption{Failure cases of our method. Some zero-shot or few-shot characters in text images still bring difficulties to our method.}
    \label{fig:failure}
\end{figure}

\paragraph{Failure cases.}
Some failure cases are shown in Figure~\ref{fig:failure}. The vertical text images with a large height-width ratio are still challenging for our method. In addition, our method also has difficulties in solving text images containing few-shot or zero-shot characters as well as previous methods. Finally, since we only consider the vertical and horizontal orientation, the performance of our method on oblique texts can be further improved. 

\section{Conclusion}
In this paper, we propose to extract orientation-independent features for Chinese text recognition by disentangling the content and orientation information. Specifically, we develop a character image reconstruction network to generate corresponding printed character images with two types of disentangled information. The proposed method surpasses previous methods on a scene dataset of Chinese text recognition. To benchmark the performance of existing methods on vertical text images, we collect a vertical Chinese text recognition dataset. Compared with state-of-the-art methods, the proposed method achieves around 45\% improvement.

\section*{Acknowledgements}
This work was supported in part by the National Natural Science Foundation of China (No.62176060), STCSM projects (No.20511100400, No.22511105000), Shanghai Municipal Science and Technology Major Project (No.2021SHZDZX0103), Shanghai Research and Innovation Functional Program (No.17DZ2260900), and the Program for Professor of Special Appointment (Eastern Scholar) at Shanghai Institutions of Higher Learning.

\bibliographystyle{named}
\bibliography{ijcai23}

\end{document}